\documentclass[11pt, a4paper, logo, copyright]{googledeepmind}

\usepackage[authoryear, sort&compress, round]{natbib}
\usepackage{amsmath,amsfonts,bm}

\def\eqref#1{equation~\ref{#1}}
\def\1{\bm{1}}

\DeclareMathAlphabet{\mathsfit}{\encodingdefault}{\sfdefault}{m}{sl}
\SetMathAlphabet{\mathsfit}{bold}{\encodingdefault}{\sfdefault}{bx}{n}

\usepackage{times}
\usepackage{latexsym}
\usepackage[T1]{fontenc}
\usepackage[utf8]{inputenc}
\usepackage{microtype}
\usepackage{inconsolata}
\usepackage{graphicx}
\usepackage{balance}       % to better equalize the last page
\usepackage{graphics}      % for EPS, load graphicx instead 
\usepackage{hyperref}
\usepackage{color}
\usepackage{booktabs}
\usepackage{textcomp}
\usepackage{subcaption}
\usepackage{enumerate}
\usepackage{xcolor}
\usepackage{lipsum}% http://ctan.org/pkg/lipsum
\usepackage{makecell}
\usepackage{multicol}
\usepackage{multirow}
\usepackage{array}
\usepackage{verbatimbox}
\usepackage{enumitem}
\usepackage{amsmath}
\usepackage{float}
\usepackage{stfloats}
\usepackage{graphicx}
\usepackage{amsthm}
\usepackage{listings}
\usepackage{caption} 
\usepackage[export]{adjustbox}
\usepackage{xspace}
\usepackage{epsfig}
\usepackage[linesnumbered]{algorithm2e}
\usepackage{algpseudocode}
\usepackage{tabularx}
\usepackage{arydshln}
\usepackage[bottom]{footmisc}
\usepackage{tcolorbox}
\usepackage{stackengine}
\usepackage{placeins}

\usepackage[normalem]{ulem}
\useunder{\uline}{\ul}{}
\tcbuselibrary{breakable}

\definecolor{E+F}{RGB}{	255, 99, 71}
\definecolor{B+F}{RGB}{255, 165, 0}
\definecolor{E+I}{RGB}{	173, 216, 230}
\definecolor{B+I}{RGB}{	30, 144, 255}
\definecolor{D}{RGB}{	60, 179, 113}

\definecolor{maroon}{cmyk}{0,0.87,0.68,0.32}

\usepackage{tikz}

\definecolor{darkgreen}{rgb}{0.0, 0.5, 0.0}
\definecolor{usercolor}{RGB}{200, 230, 250} % Lighter pastel blue
\definecolor{coachcolor}{RGB}{180, 250, 180} % Lighter pastel green

\title{Use of Continuous Glucose Monitoring with Machine Learning to Identify Metabolic Subphenotypes and Inform Precision Lifestyle Changes}
\reportnumber{} % Leave blank if n/a
\renewcommand{\today}{2025-10-30}

\author[1,2,3]{Ahmed A. Metwally, PhD}
\author[4,$\ddagger$]{Heyjun Park, PhD}
\author[1,$\ddagger$]{Yue Wu, PhD}
\author[5,$\dagger$]{Tracey McLaughlin, MD}
\author[1,$\dagger$]{Michael P. Snyder, PhD}

\affil[1]{Department of Genetics, Stanford University, Stanford, CA, USA}
\affil[2]{Google Research, Mountain View, CA, USA}
\affil[3]{Systems and Biomedical Engineering Department, Cairo University, Giza, Egypt}
\affil[4]{Department of International Health, Johns Hopkins Bloomberg School of Public Health, Baltimore, MD, USA}
\affil[5]{Department of Medicine, Stanford University, Stanford, CA, USA}
\affil[$\ddagger$]{Equal contribution}
\affil[$\dagger$]{Correspondence: tmclaugh@stanford.edu, mpsnyder@stanford.edu}

\begin{abstract}

\section*{Abstract}
The classification of diabetes and prediabetes by static glucose thresholds obscures the pathophysiological dysglycemia heterogeneity, primarily driven by insulin resistance (IR), $\beta$-cell dysfunction, and incretin deficiency. This review demonstrates that continuous glucose monitoring and wearable technologies enable a paradigm shift towards non-invasive, dynamic metabolic phenotyping. We show evidence that machine learning models can leverage high-resolution glucose data from at-home, CGM-enabled oral glucose tolerance tests to accurately predict gold-standard measures of muscle IR and $\beta$-cell function. This personalized characterization extends to real-world nutrition, where an individual’s unique postprandial glycemic response (PPGR) to standardized meals, such as the relative glucose spike to potatoes versus grapes, could serve as a biomarker for their metabolic subtype. Moreover, integrating wearable data reveals that habitual diet, sleep, and physical activity patterns, particularly their timing, are uniquely associated with specific metabolic dysfunctions, informing precision lifestyle interventions. The efficacy of dietary mitigators in attenuating PPGR is also shown to be phenotype-dependent. Collectively, this evidence demonstrates that CGM can deconstruct the complexity of early dysglycemia into distinct, actionable subphenotypes. This approach moves beyond simple glycemic control, paving the way for targeted nutritional, behavioral, and pharmacological strategies tailored to an individual’s core metabolic defects, thereby paving the way for a new era of precision diabetes prevention.

\end{abstract}
\begin{document}

\maketitle
% ~\citep{shah2025normal}
\section{Introduction}

\subsection{Significance of Diabetes and Need to Define Heterogeneity} Diabetes affects 38.1 million, or 14.7\%, of all U.S. adults aged 18 years or older, and another 38 million, or 38\%, have prediabetes, of whom less than 20\% are aware~\citep{CDC2024}. Diabetes and prediabetes are classified based on the degree of glucose elevation, but this does not reflect the multiple physiological pathways that can lead to elevations in blood glucose. Specifically, insulin resistance (IR), beta-cell failure, incretin deficiency, and unrestrained hepatic glucose production (hepatic IR) all contribute. If it were possible to determine which process/processes were the predominant contributors to dysglycemia in a given individual, targeted therapy could be employed, potentially yielding more successful diabetes prevention and treatment than following a one-size-fits-all approach. Historically, both lifestyle and medical interventions have been recommended based on the degree of overweight and glucose elevation without regard for 1) interindividual variability in pathophysiology; 2) risk trajectory, or 3) anticipated individualized treatment response. With new technologies such as continuous glucose monitoring, it may be possible to characterize and subtype early stages of dysglycemia that reflect/predict these important characteristics, thus significantly impacting human health. In the following review, we outline several novel aspects of CGM and other wearable devices that make inroads into better characterizing metabolic heterogeneity of adults with early dysglycemia, and highlight the interactions between metabolic physiology and response to specific lifestyle interventions. 

\subsection{Limitations of Current Diabetes Classification} Diabetes and prediabetes are diagnosed based on glucose cutoffs~\citep{ADA2025, Qu2022, Noga2024, Shen2025}. Thus, diagnosis is based on an elevated plasma glucose without regard for the mechanism that led to the elevation. We now know that many pathophysiological pathways contribute to glucose elevation – and yet these are lumped together under the umbrella of diabetes. Current diabetes classification of non-pregnant adults is limited to type 1, type 2, and other specific forms~\citep{ADA2025, Qu2022, Noga2024, Shen2025}. Of all prevalent cases of diabetes, type 2 represents > 95\%. It is becoming clear that not all type 2 diabetes patients follow the same clinical trajectory, with some more prone than others to developing deterioration of glucose control, cardiovascular events, nephropathy, retinopathy, or fatty liver disease~\citep{Hulman2017, Kobayashi2025, Karpati2018, Anjana2020}. Similarly, it is also known that not all patients respond equally well to the same diabetes medications~\citep{Nair2022}. Thus, there is a movement towards subclassifying type 2 diabetes to better reflect the heterogeneity and inform targeted treatments. Methods for subtyping diabetes/prediabetes range from genetic~\citep{Udler2019} to clinical/laboratory biomarkers~\citep{Ahlqvist2018}. We recently proposed that CGM represents a novel and accurate method that can be used to subtype the early dysglycemic state according to underlying metabolic physiology~\citep{Metwally2024}. This method has the advantage over genetic subclassification of reflecting the current clinical state which can change as a result of contributory environmental factors (e.g., body weight, physical activity), and over both genetic and clinical/demographic factors in degree of accuracy~\citep{Metwally2024}. It does not require laboratory tests or measurement of waist circumference, which pose potential obstacles. Furthermore, because only a single CGM sensor is needed, the cost and burden are minimal, and because it does not require access to a health care system, it bypasses socioeconomic and geographic barriers that might preclude diagnosis/subphenotyping in a standard health care system that is relatively inaccessible to low income and/or geographically remote populations. Lastly, this method is amenable to tracking changes or stability in metabolic health over time, as demonstrated in the relative stability of a five-point OGTT glucose trajectory over three years~\citep{Hulman2018}. 
 
\subsection{Overview of why postprandial glucose pattern would be good for metabolic phenotyping based on dynamic features of the glucose curve} Glucose elevations following an oral nutrient challenge have been used in clinical practice since the development of the first oral glucose tolerance test (OGTT) in 1917~\citep{Jagannathan2020}. While the use of this test has changed over time, early in its evolution, post-glucola metrics including the 1-hour glucose, the integrated area-under-the-curve glucose, whether the shape of the glucose curve from a five point test (0, 30, 60, 90, 120 min) was mono- or bi-phasic or unclassified was used to identify both underlying metabolic features of IR and beta-cell dysfunction, as well as individuals at highest risk of progressing to diabetes~\citep{Cheng2015, AbdulGhani2008, Alyass2015, Kim2016, Tanczer2020, Kim2012, Kanauchi2005, Bervoets2015, Chung2017, Kasturi2019, Cheng2019, Kaga2020, Manco2017, Ismail2018, AbdulGhani2010}. Later, mathematical approaches such as functional component analysis~\citep{Froslie2013}, and latent trajectory analysis~\citep{Hulman2018} were explored as biomarkers of physiologic subtypes of diabetes. Due to the burden of the five-point OGTT, however, this was soon abandoned in favor of the two-point (0 and 120 min) test with use of values for both fasting and 120 min as diagnostic thresholds for diabetes based on the correlation with development of retinopathy~\citep{ExpertCommittee1997, Hanson1993, Engelgau1997}, and use of more complex algorithms or “shape of the curve” glucose for subphenotyping diabetes or prediabetes was not pursued. 

With the advent of CGM, the availability of a more nuanced shape of the curve merits deeper exploration of glucose patterns as a potential subphenotyping strategy that reflects both underlying physiology and risk of progressing to diabetes and related metabolic complications such as cardiovascular disease.

\section{Predicting Metabolic Subphenotypes with OGTT Glucose Time Series and Machine Learning Approach}

\subsection{Predicting metabolic subphenotypes using glucose time-series data and machine learning approach from frequently sampled OGTT}
\begin{figure}[htbp]
    \centering
    \includegraphics[width=0.88\textwidth]{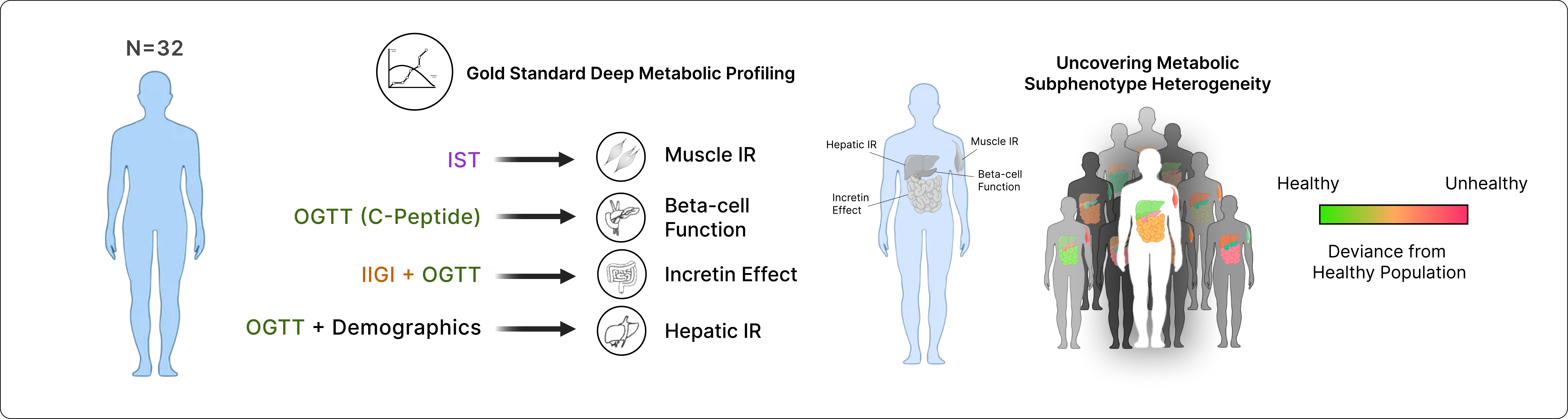}
    \begin{minipage}{\textwidth}
        \caption{Deep metabolic profiling via gold-standard quantitative tests were used in 13 to evaluate: (1) Muscle insulin resistance via the Insulin Suppression Test (IST), (2) Beta-cell function by calculating the disposition index from an Oral Glucose Tolerance Test (OGTT) using C-peptide deconvolution, (3) Incretin effect using an Isoglycemic Intravenous Glucose Infusion (IIGI) and an OGTT, (4) Hepatic insulin resistance using a validated formula based on OGTT insulin levels and demographic information.}
        \label{fig:figure1}
    \end{minipage}
\end{figure}

Initial efforts to identify metabolic subphenotypes from the glucose shape of the cure utilized a five-point OGTT latent class trajectory analysis in 5,861 patients without diabetes from the Danish Inter99 cohort~\citep{Hulman2017}. In this approach, patients underwent metabolic characterization using the euglycemic clamp for insulin sensitivity (M/I) and the IVGTT for insulin secretion (GSIR) with a calculated disposition index (GSIR * M/I) to measure beta cell function; glucose shape of the curve was determined from plasma at time 0, 30, 60, 90, and 120 min during standardized 75g, 2-hour OGTT. Four different glucose patterns were identified and found to predict both baseline metabolic profile and subsequent clinical risk for diabetes, retinopathy, nephropathy, cardiovascular disease, and all-cause mortality over a 13 year follow up period~\citep{Hulman2017}. In a separate study, these curve profiles were also evaluated for stability over time: individuals subgroup based on glucose curve remained the same in 55\% over three years; moved to a more favorable profile in 19\%, and to a less favorable profile in 27\%, which was associated with increase in BMI and related metabolic risk markers~\citep{Metwally2024}. In this study, metabolic physiologic profile was deeply phenotyped using gold-standard metabolic tests in a cohort of individuals with normoglycemia or prediabetes (Figure~\ref{fig:figure1}). These rigorous tests precisely quantified four key physiological processes: muscle IR (measured by modified insulin-suppression test (SSPG), hepatic IR (inferred by a validated index), $\beta$-cell dysfunction (insulin secretion rate (ISR) quantified with c-peptide deconvolution during a 3-hour OGTT with adjustment for IR via calculation of the disposition index (ISR/ SSPG), impaired incretin function quantified by calculating relative insulin secretion during OGTT vs isoglycemic intravenous glucose infusion (IIGI)~\citep{Nauck1986}. The study revealed that individuals exhibited diverse combinations of these defects, with all but 9\% exhibiting a single dominant or co-dominant subphenotype: muscle and hepatic IR were highly correlated, accounting for single or co-dominant metabolic phenotype in 35\% of individuals, while $\beta$-cell dysfunction and/or incretin deficiency accounted for 42\%. Importantly, these underlying metabolic dysfunctions did not correlate with traditional glycemic measures like HbA1c, highlighting the inadequacy of current diagnostic approaches to subclassify early stages of dysglycemia.

\begin{figure}[htbp]
    \centering
    \includegraphics[width=0.88\textwidth]{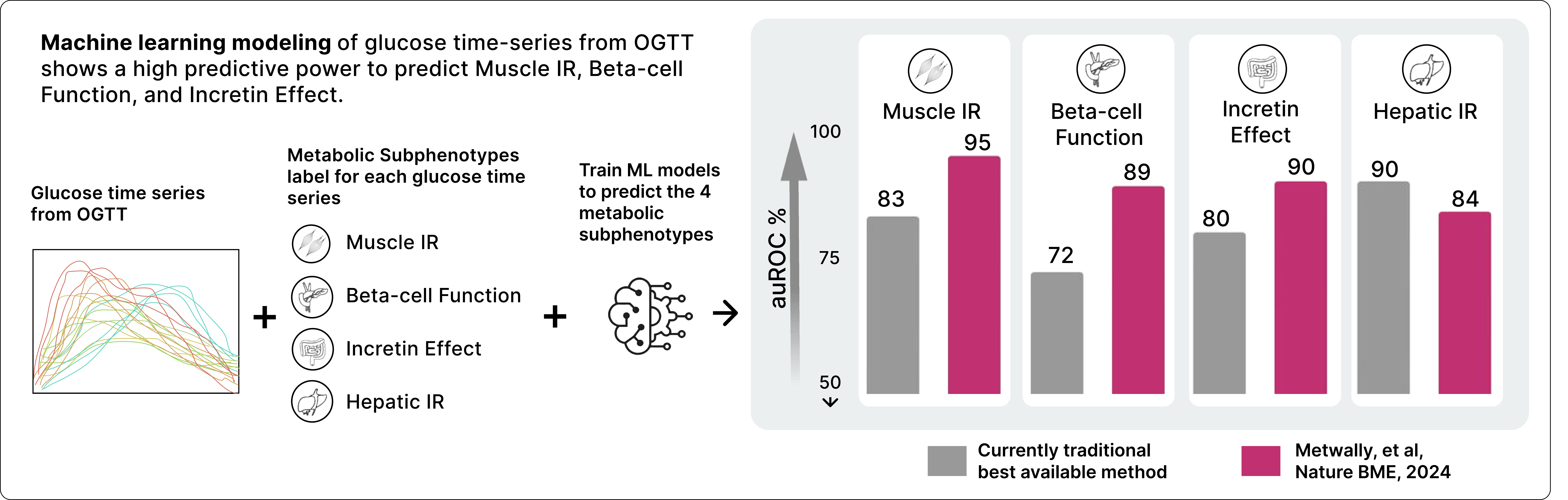}
    \begin{minipage}{\textwidth}
        \caption{Robust performance in predicting muscle IR, beta-cell dysfunction, and incretin effect using glucose timeseries compared to currently-used surrogate markers.}
        \label{fig:figure2}
    \end{minipage}
\end{figure}

To translate these complex physiological insights into a scalable diagnostic tool, the team developed a machine-learning ML framework that utilizes features extracted from a frequently sampled, 16-point plasma glucose curve obtained during OGTTs in the CTRU. This dense sampling effectively mimicked the continuous data stream provided by a CGM, offering a "CGM-like" assessment in the research setting. Two main feature extraction approaches were employed: "OGTT\_G\_Features," encompassing 14 engineered features (e.g., AUC, peak glucose, slopes), and "OGTT\_G\_ReducedRep," a reduced representation using principal component analysis. This approach demonstrated the superior predictive power of ML models trained on high-resolution plasma glucose data for predicting metabolic subphenotypes compared to current clinical biomarkers, laboratory tests, and even polygenic risk scores~\citep{Mahajan2022, Metwally2024}. Specifically, muscle IR was predicted with a robust area under the receiver operating characteristic curve (auROC) of 0.95 using OGTT\_G\_ReducedRep, as compared to HOMA-IR with auROC of 0.77, while $\beta$-cell dysfunction and incretin deficiency achieved auROCs of 0.89 and 0.88, respectively, which compared to auROCs of 0.48 and 0.68 for HOMA-ß and 1 hour GLP/GIP, respectively (Figure~\ref{fig:figure2}). While prediction for hepatic IR was promising (0.84 AUC), other surrogate markers, such as HOMA-IR~\citep{Matthews1985} and Matsuda index~\citep{Matsuda1999}, showed comparable or superior performance. 

This work demonstrated accuracy in identifying metabolic subphenotypes from the shape of the glucose curve, derived from highly sampled plasma OGTTs, represents a crucial step. It validates the concept that dynamic glucose responses, particularly when captured with high fidelity akin to CGM data, contain rich information about underlying metabolic physiology that standard, less frequent measurements miss. The robust performance of these models, initially trained and validated using research-unit plasma data, sets the foundation for a practical, non-invasive method for early risk stratification and targeted intervention for individuals at risk of T2D, ultimately enhancing precision approaches to diabetes management.

\subsection{Predicting metabolic subphenotypes from CGM during at-home OGTT} 
\begin{figure}[htbp]
    \centering
    \includegraphics[width=0.88\textwidth]{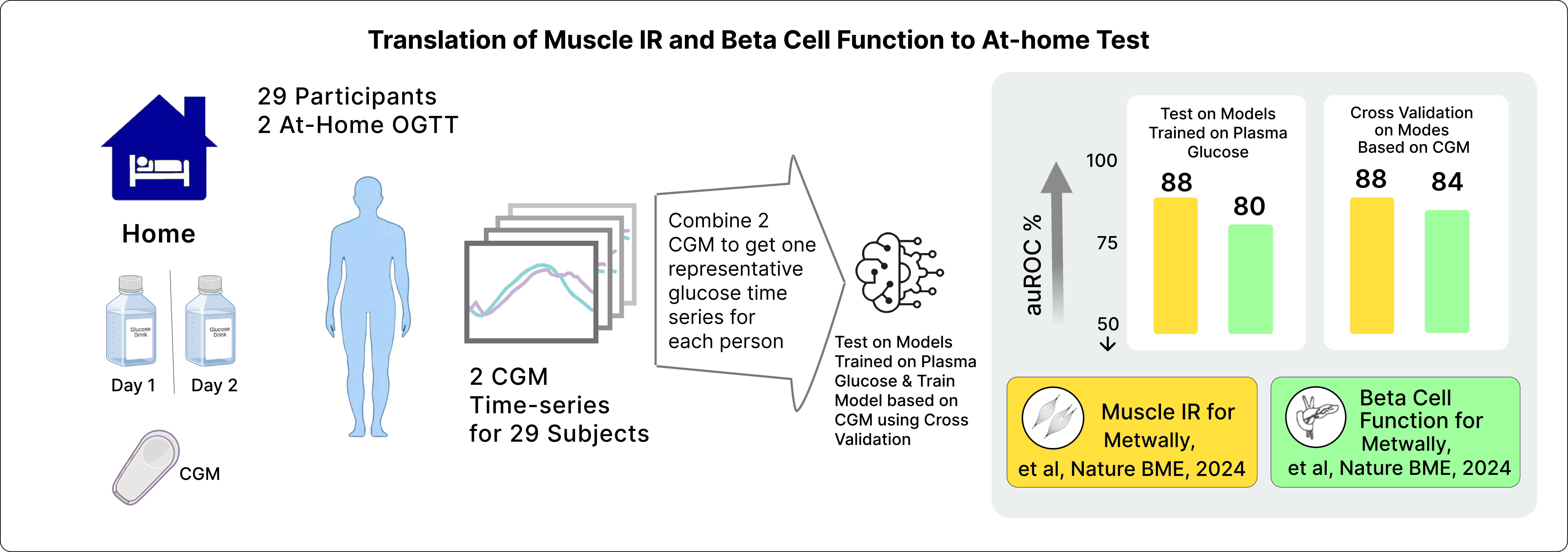}
    \begin{minipage}{\textwidth}
        \caption{Study design of the at-home OGTT test via CGM to predict muscle IR and $\beta$-cell function. Participants underwent gold-standard testing at the research unit for insulin resistance (Insulin Suppression Test) and $\beta$-cell function (16-point OGTT with C-peptide deconvolution adjusted for SSPG and expressed as DI) as described, as well as two OGTTs administered at home under standardized conditions during which glucose patterns were captured by a CGM within a single 10-day session (Dexcom G6 pro).}
        \label{fig:figure3}
    \end{minipage}
\end{figure}

To assess the feasibility of using at-home CGM testing to predict muscle IR and $\beta$-cell function, Metwally et al. recruited 29 participants who underwent both gold-standard clinical tests (the SSPG test for IR and a 16-point OGTT with C-peptide deconvolution for $\beta$-cell function) and two standardized OGTTs at home while their glucose levels were recorded by a CGM (Dexcom G6 pro) (Figure~\ref{fig:figure3}).

To ensure reproducibility, scalability, and minimize environmental influences, participants were instructed to perform the two at-home OGTTs within a single 10-day CGM sensor session after an overnight fast. They consumed a 75-g glucola drink identical to the in-clinic procedure, remaining seated for 3 hours with no intake other than water. Detailed instructions were provided regarding diet and physical activity before the test, avoiding celebratory or restrictive eating, strenuous exercise, and late-night meals. This meticulous standardization aimed to control variability, a known challenge with OGTTs.
 
A crucial aspect of feasibility was demonstrating that at-home CGM data were comparable to in-clinic, plasma-based measurements. The study found high concordance between glucose time-series data obtained from (1) In-clinic plasma sampling and simultaneous in-clinic CGM measurements (Pearson correlation of 0.81), (2) two repeated at-home OGTTs using CGM (Pearson correlation coefficient r=0.86), indicating strong reproducibility of the at-home method, and (3) In-clinic venous measurements and the mean of two at-home CGM OGTTs (r=0.80). These high correlations indicate that CGM measurements performed at home are similar to those obtained in a clinical research unit and are highly reproducible when conducted under standardized conditions.
 
The machine-learning framework developed in the first part of the study (using frequently sampled glucose timeseries from OGTT done at the research unit), when applied to CGM-generated glucose curves from at-home OGTTs, accurately predicted metabolic subphenotypes. IR was predicted with an AUC of 88\%. Similar to the plasma-based OGTT at research unit, at-home CGM curves from at-home OGTT clearly showed differences between insulin-resistant and insulin-sensitive individuals, with IR individuals exhibiting higher glucose peaks, broader curves, and higher AUCs. Moreover, $\beta$-cell dysfunction was predicted with an AUC of 80\% using the mean of two home CGM curves. This performance further improved to an AUC of 84\% when using cross-validation based solely on CGM data.

Results showed that classification based on the mean of two home CGM curves performed even better than the classification based on plasma glucose curves from the independent test set (auROC of 0.88 vs 0.72 for IR, and auROC of 0.8 vs 0.76 for $\beta$-cell dysfunction). This improved performance is attributed to the efficient information extraction from continuous CGM measurements, which provide much higher frequency data than less frequent venous sampling.
 
While historical OGTTs have noted intra-individual variability, the study's standardized approach for at-home CGM resulted in a low coefficient of variation (11\% for CGM across all timepoints), which compares favorably to published variability for 2-hour plasma glucose (16.7\%). This suggests that proper patient instructions can significantly minimize environmental stressors leading to glucose response variability.
 
In summary, this novel approach demonstrates the feasibility of conducting an at-home OGTT with a CGM, which provides a practical, accurate, and scalable method for identifying distinct metabolic subphenotypes of T2D, such as IR and beta-cell function. It removes the patient burden associated with multiple blood draws over a 3-hour window in a clinical setting, making the OGTT more feasible and accessible for widespread use. By leveraging high-resolution glucose data from at-home CGM-enabled OGTTs and machine learning, the study demonstrates a significant step towards enabling precision medicine for diabetes prevention and treatment, allowing for targeted therapies based on an individual's specific underlying metabolic defects.

\section{Predicting Metabolic Subphenotypes Using CGM at Home with Real Food Challenges}
\begin{figure}[htbp]
    \centering
    \includegraphics[width=0.88\textwidth]{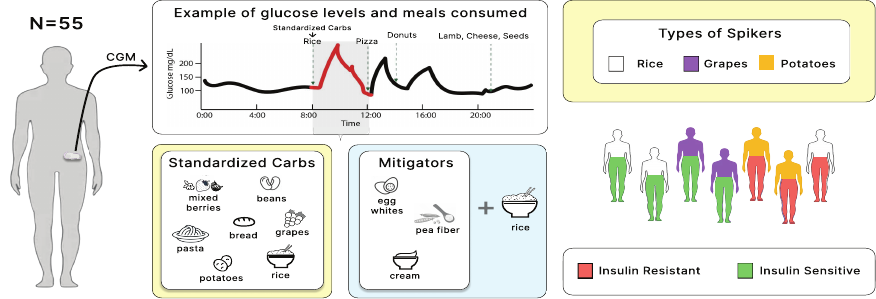}
    \begin{minipage}{\textwidth}
        \caption{Study design of the standardized meal study. 55 participants consumed 7 different standardized carbohydrate meals and 3 mitigator foods with rice with replicates while wearing CGM. Participants were grouped by their highest response to carbohydrate and measured metabolic subphenotypes.}
        \label{fig:figure4}
    \end{minipage}
\end{figure}

Variation in individuals’ postprandial glycemic response (PPGR) to the same meals has been clearly demonstrated~\citep{Zeevi2015, Berry2020}. The determinants of this variability are incompletely characterized, but several studies have implicated microbiome, genetics, meal composition and timing, and other patient characteristics as contributors, and models to predict PPGR based on these inputs have shown reasonable success (r= 0.68 to 0.77)~\citep{Zeevi2015, Hall2018, Wu2025, Berry2020} 39,41,4239,40. A subsequent study from our group demonstrated that inter-individual variability on PPGR was a function of underlying metabolic phenotype~\citep{Wu2025}. In this study, 55 participants without diabetes were recruited for deep metabolic subphenotyping as measured by gold-standard tests in the metabolic research unit, as per above. Participants consumed seven different carbohydrate meals in replicate with the same amount of total carbohydrates (50 g) on different days in the morning after an overnight fast and under standardized conditions to remove environmental confounders (e.g., coffee, exercise, late night snack, eating out the night prior) while wearing a Dexcom CGM. The seven carbohydrate meals included rice, bread, potatoes, pasta, black beans, mixed berries (blackberries, strawberries, and blueberries), and grapes. Eleven time-series features were extracted from the CGM data to quantify PPGR, with a focus on the 3 hours after meals, such as AUC(>baseline), delta glucose peak (the difference between the peak and baseline glucose), and the time from baseline to the peak. Clinical data were also collected from participants at baseline visits (Figure~\ref{fig:figure4}).

\begin{figure}[htbp]
% \begin{figure}[ht]
    \centering
    \includegraphics[width=0.88\textwidth]{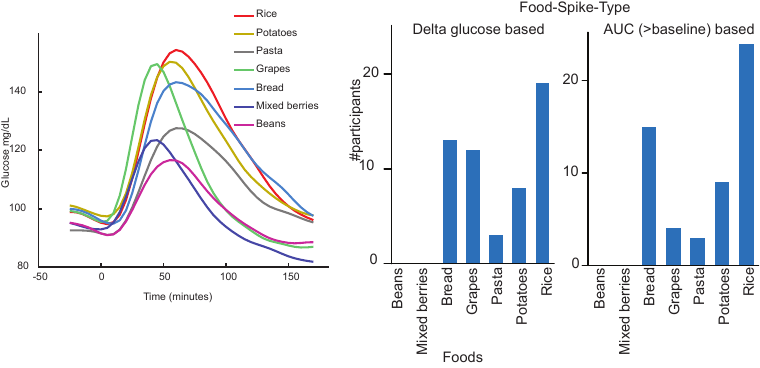}
    \begin{minipage}{\textwidth}
        \caption{Postprandial glycemic responses to different carbohydrates. Left: Mean CGM curves after different meals. The X-axis is time, with the food log consumption time as 0, and the Y-axis is glucose level. Right: Number of participants classified to each spiker type as defined by both delta glucose peak and AUC(>baseline). The X-axis indicates different carbohydrates, and the Y-axis is the number of participants for whom a given carbohydrate produced the highest spike.}
        \label{fig:figure5}
    \end{minipage}
\end{figure}

The researchers found that different participants had different responses to the same carbohydrate meals and had different carbohydrates producing the highest PPGRs as measured by AUC (>baseline) and delta glucose~\citep{Wu2025}. Participants were stratified into ‘spiker’ type based on which carbohydrate meal produced the highest average PPGR. For instance, when grouped according to delta glucose peak, rice-spikers were the largest group (35\%), followed by bread-spikers (24\%) and grape-spikers (22\%) (Figure~\ref{fig:figure5}).

\begin{figure}[htbp]
    \centering
    \includegraphics[width=0.88\textwidth]{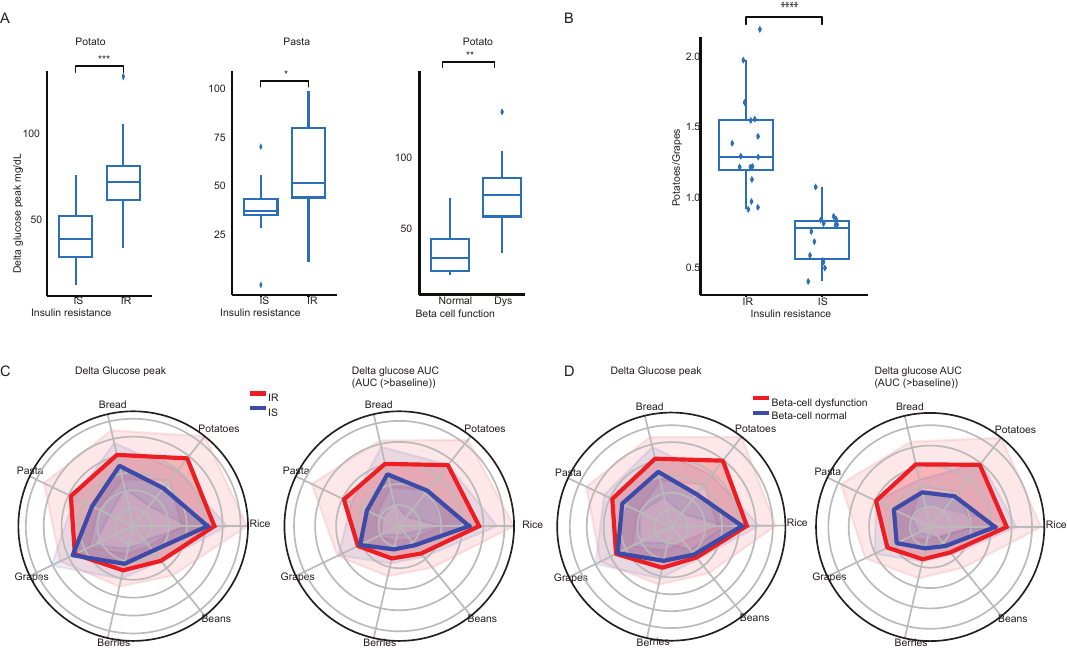}
    \begin{minipage}{\textwidth}
        \caption{Differential responses to potato and pasta across people with different metabolic subphenotypes. A: Delta glucose peaks after eating potatoes and pasta are compared between groups with IR and IS, and delta glucose peaks after eating potatoes are compared between groups with normal and dysfunctional (dys) beta cell functions. B: The ratio between the  response (delta glucose) to potato and that to grapes are compared between groups with IR and IS. C: Average delta glucose peak and AUC(>baseline) between participants with IR and IS. D, Average delta glucose and AUC(>baseline) between participants with different beta cell functions (normal and dysfunctional). The solid line is the average value, and the shade is one s.d.}
        \label{fig:figure6}
    \end{minipage}
\end{figure}

Interestingly, PPGR to the same foods differed significantly according to metabolic subphenotype. The PPGR to potatoes and pasta, but not other carbohydrates, differed significantly as a function of IR, and the PPGR to potatoes differed significantly in those with normal or insufficient beta-cell function (Figure~\ref{fig:figure6}A). The relative elevation in PPGR to potato in comparison to grapes (potato-grape ratio) differed significantly between the metabolically healthy (insulin sensitive) and unhealthy (insulin resistant), with almost no overlap, suggesting it might be used as a biomarker for metabolic subphenotype (Figure~\ref{fig:figure6}B). While more work will be needed to validate this biomarker in larger populations, it is the first time a glycemic response to real food has been demonstrated to predict underlying metabolic physiology. Of note, not only were “potato spikers” more likely to be insulin resistant and/or beta-cell deficient, they also exhibited higher HbA1 and fasting blood glucose, indicating a worse metabolic state, in particular, compared with grape-spikers. Figures~\ref{fig:figure6}C and ~\ref{fig:figure6}D illustrate how the relative elevations in delta glucose peak and AUC, derived from all tested standardized meals, relate to insulin resistance and beta-cell function.

The researchers~\citep{Wu2025} further tested the potential biochemical mechanisms of the connection between the PPGR to potatoes and pasta with metabolic subphenotypes. Biochemical assays comparing potato, pasta, bread, and rice meals to measure the content of different starch components after following the cooking instructions. Components measured included resistant starch (which is a fiber), rapidly digestible starch, and slowly digestible starch. Potatoes, which in this study were hash browns that were cooked and then cooled, had significantly higher resistant starch, and pasta had significantly higher slowly digestible starch than bread and rice.

The results of this study demonstrate for the first time an association between metabolic subphenotype and an individual’s relative PPGR to some carbohydrates over others. This may have implications for both personalized clinical nutrition and diabetes prevention (avoid potatoes if you are insulin resistant or beta-cell deficient) but also in risk stratification by identifying individuals at highest risk for progression to diabetes, who would benefit from a variety of intentions that are known to reduce the risk of diabetes that extend beyond avoiding potatoes - for example weight loss, physical activity, metformin, and a variety of other medications have all been shown to prevent diabetes in high risk individuals~\citep{Knowler2002}. Leveraging CGM to further explore differential PPGR responses as predictors of metabolic subphenotypes merits additional study. 

\section{Associations between metabolic subphenotypes and real world lifestyle patterns (dietary composition, sleep, physical activity) measured by wearable devices}

\subsection{Lifestyle's role in metabolic subphenotypes} 
Diet, sleep, and physical activity are core lifestyle behaviors that can be modified to support metabolic health~\citep{Teymoori2023, Dwibedi2022, Legaard2023, Chatterjee2018}. Their effects on glucose regulation have been widely studied, primarily in relation to glucose levels~\citep{Knowler2002}. Specifically, certain dietary components (e.g., nutrients and food groups) and broader dietary patterns are known to influence fasting blood glucose, hemoglobin HbA1c, and IR~\citep{Berry2020, Asnicar2021, Bermingham2024}. Similarly, both sleep duration and quality, along with physical activity levels, have demonstrated associations with these glycemic markers~\citep{ADA2025, Qu2022, Noga2024, Shen2025}. However, important gaps remain in our understanding of how these behaviors contribute to the development of metabolic diseases. Notably, few studies have assessed all three lifestyle domains in an integrated manner. Furthermore, no study has examined how these behaviors relate to underlying pathophysiological processes of type 2 diabetes, including beta-cell dysfunction, IR, and impaired incretin response. Although these metabolic impairments often arise before overt glycemic dysregulation, their connections to habitual lifestyle patterns remain poorly defined. Therefore, more studies are needed to investigate relationships that could clarify mechanisms of early disease development and help identify more precise behavioral intervention targets.

\begin{figure}[htbp]
    \centering
    \includegraphics[width=\textwidth]{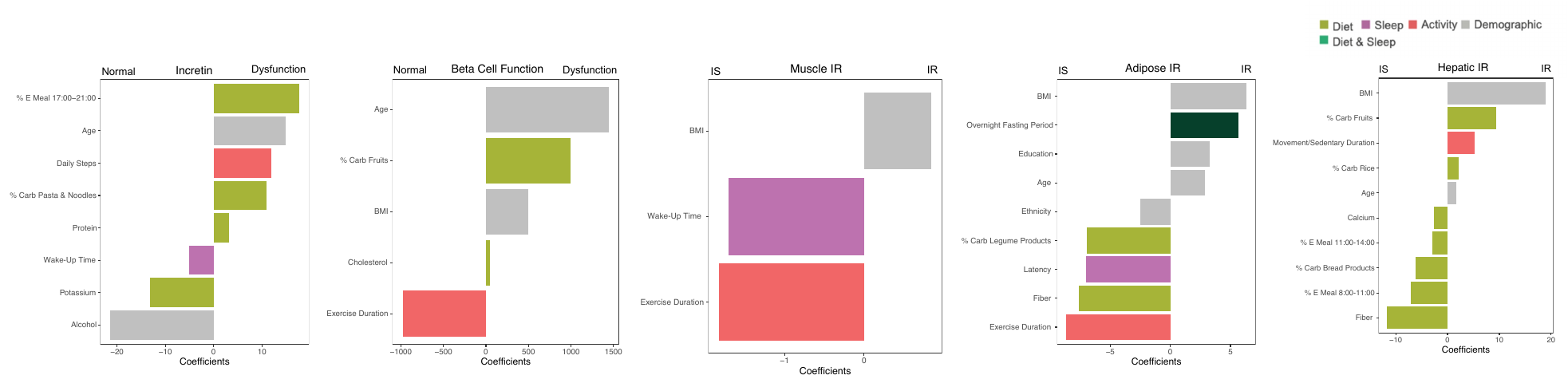}
    \begin{minipage}{\textwidth}
        \caption{Integrated lifestyle prediction models predicting metabolic subphenotypes. LASSO classification models were built using all lifestyle features, and the coefficients of the final predictors are shown. Classifications include incretin and beta-cell normal function vs. dysfunction as well as muscle, adipose, and hepatic insulin sensitivity (IS) vs. resistance (IR). Colors denote feature categories. Sex (1 = male, 0 = female) and ethnicity (1 = Caucasian, 0 = non-Caucasian) are numeric variables. Latency is the time from wakefulness to sleep onset. \%E Meal and \%Carb indicate proportional energy or carbohydrate contribution. Movement/sedentary ratio and education (years) are continuous variables. Nutrients (e.g., fiber, sodium) represent daily intakes.}
        \label{fig:figure7}
    \end{minipage}
\end{figure}

Park et al. extended this research by examining beyond the conventional associations between lifestyle behaviors and glycemic markers to investigate how these behaviors relate to distinct metabolic characteristics (herein, “metabolic subphenotypes”)~\citep{Park2025}. They hypothesized that lifestyle factors would be associated not only with standard glycemic outcomes such as fasting glucose, A1C, and oral glucose tolerance test (OGTT) results, but also with specific physiological dysfunctions, including IR, beta-cell dysfunction, and compromised incretin response. A mobile food logging application was used for dietary data collection, and sleep and physical activity data were collected through wearable devices. The extensive multimodal data from three lifestyle domains, along with demographic covariates, were integrated into predictive models for distinct metabolic subphenotypes. Notably, they identified unique sets of lifestyle predictors in each model (Figure~\ref{fig:figure7}). Age and BMI were shared predictors across several dysfunctional metabolic subphenotypes, such as incretin dysfunction, beta-cell dysfunction, and IR in all specific tissues (muscle, adipose tissue, and liver). Among lifestyle factors, total exercise duration predicted normal beta-cell function as well as insulin sensitivity in muscle and adipose tissues. However, it was not a significant predictor of hepatic insulin sensitivity. Distinct sleep characteristics were also differentially associated with subphenotypes: wake-up time was associated with muscle IR and incretin function, whereas sleep latency (i.e., the time to transition from full wakefulness to sleep onset) was related to IR in adipose tissue. 

In the dietary domain, fiber intake showed protective associations with both adipose and hepatic IR, but no dietary factor significantly predicted muscle IR. Interestingly, hepatic insulin sensitivity appeared to be mostly associated with dietary factors, including not only dietary composition but also meal timing. These findings highlight the complexity of metabolic regulation and its variation across tissues. This work also emphasizes the potential of prevention strategies that are tailored to an individual’s habitual lifestyles and metabolic susceptibility. A precision-based approach may offer more effective results in preventing the onset of type 2 diabetes.

\subsection{CGM to identify the optimal timing of lifestyle behaviors by metabolic subphenotypes}
A body of growing evidence has shown that lifestyle behaviors act as circadian cues that impact numerous genomic and metabolic processes, including glucose regulation~\citep{Panda2016, Tahara2013, Wehrens2017}. The timing of introducing lifestyle behaviors (e.g., eating, physical activity, and sleep) can either support or disrupt the body’s intrinsic metabolic rhythms. When these behaviors are misaligned with the endogenous circadian clock, they may impair normal metabolic regulation and increase the risk for cardiometabolic diseases~\citep{Scheer2009, Nakamura2021, Shen2023, Pan2011, Gabriel2019, McHill2017}. In contrast, aligning daily routines with intrinsic circadian timing may enhance efficiency and promote optimal metabolic health~\citep{Chaix2014, Pavlou2023}.

Recent advances in wearable technologies have significantly transformed research in this area by collecting high-frequency, time-stamped, and dense physiological data~\citep{AncoliIsrael2003, Shajari2023, Ohkawara2011, Zambotti2019}. Modern wearable devices can capture thousands of data points per participant at intervals as frequent as every 1 to 5 minutes. This high-resolution of the data allows researchers to characterize individual’s behavioral profiles with remarkable precision at scale. In contrast to traditional methods based on self-reported surveys or infrequent sporadic measurements, diverse physiologic metrics measured via wearables offer continuous and dynamic view of lifestyle patterns, including how the timing and regularity of behaviors are linked to metabolic outcomes. 

Evidence from Park et al. supports the importance of the timing of all three lifestyle behaviors in glycemic regulation detected from CGM. In the dietary domain, higher energy intake between 5:00 and 9:00 p.m. was consistently associated with poorer nighttime glucose control, including less time spent in time-in-range (TIR; 70-100 mg/dL), more time in nighttime hyperglycemia (>100 mg/dL), and elevated mean glucose levels the following day. Irregular sleep timing, particularly high variability in bedtime across days, was also associated with higher maximum glucose levels the next day. Time-series analyses integrating step counts and CGM data showed divergent temporal patterns of glycemic response (captured by CGM) between individuals with insulin sensitivity (IS) and IR. In the IS group, an increase in physical activity between 2:00 and 5:00 p.m. conferred glycemic benefits lasting up to 48 hours. In contrast, among the IR group, morning activity (8:00 to 11:00 a.m.) was associated with improvements lasting over 24 hours.

Together, these findings underscore the importance of not just what individuals do, but when they do it. Integrating circadian principles into lifestyle interventions and leveraging wearable data to tailor timing strategies may offer new opportunities to improve metabolic health. Future research should focus on mechanistic studies and targeted interventions that explicitly examine whether synchronizing lifestyle behaviors across multiple domains with internal biological clocks can help prevent or mitigate metabolic dysregulation, particularly in populations at elevated risk.

\section{Leveraging CGM to evaluate dietary mitigators of PPGR}
High PPGR is a risk factor for CVD and all-cause mortality independent of fasting blood glucose and HbA1c~\citep{Cavalot2011, Gallwitz2009}. Therefore, reducing PPGR to meals (mitigation) is one of the goals of nutrient intervention. In Wu et al., in addition to the 7 standardized carbohydrate meals, 3 mitigators containing fiber (10 g, via pea fiber), protein (10 g, via egg white), and fat (15 g, via cream) were also tested for their ability to attenuate the PPGRs to rice, which on average produced the greatest glucose elevation in their study~\citep{Wu2025}. Thirty-two participants consumed fiber, protein, or fat mitigators 10 min before consuming rice (50 g total carbohydrates). The mitigation effect was calculated as a normalized difference between PPGRs of rice $+$ mitigator and rice. A negative value indicated that consuming the mitigator in front reduced the PPGR compared to rice alone.

\begin{figure}[htbp]
    \centering
    \includegraphics[width=0.88\textwidth]{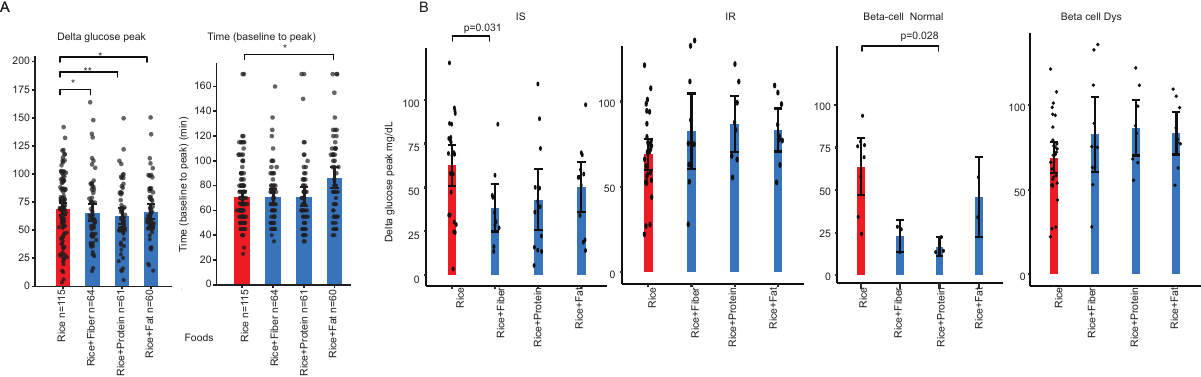}
    \begin{minipage}{\textwidth}
        \caption{ Mitigator effects of fiber, protein and fat in different groups. A: the effect of mitigators on delta glucose and time from baseline to peak in the whole cohort. B: The effect of mitigators on delta glucose on the group with IR, IS, normal beta cell, and dysfunctional beta cell. }
        \label{fig:figure8}
    \end{minipage}
\end{figure}

Across the entire cohort, all three mitigators produced a modest but statistically significant reduction in PPGR to rice. Fat, but not protein or fiber, delayed the time from the baseline to the peak glucose (Figure~\ref{fig:figure8}). Interindividual variability was also observed in mitigation effects. Additionally, similar associations between PPGR and metabolic subphenotypes were also observed in the mitigation effect. Individuals with IR presented minimal PPGR reduction with mitigators, whereas individuals with IS exhibited stronger responses to mitigators, with fiber reducing delta glucose significantly. Individuals with normal beta-cell function also had a protein significantly reducing the delta glucose peak.

% The authors further proposed a hypothesis linking metabolic subphenotypes and individual’s response to carbohydrates and mitigators. They suggested that the two main pathways of mitigation effect worked in synergy. The first pathway is reducing carbohydrate absorption, and the second is the whole process of incretin induction, insulin secretion (beta cell function), and insulin action on different organs. In healthy individuals, the two pathways work in synergy, and mitigators greatly reduce PPGRs. However, in people with IR or dysfunctional beta cells, the reduced absorption alone may be insufficient without the synergistic contribution of insulin function. As resistant starch is a type of fiber, the link between metabolic subphenotypes and differential PPGR to potatoes and pasta can also be explained by the same hypothesis.

The results in this study not only suggest targeted dietary intervention for individuals with different metabolic subtypes but also suggest the need of systematic deep profiling of the digestion and absorption process. In particular, individuals who are insulin resistant or have beta-cell dysfunction might require additional personalized mitigation solutions rather than the conventional dietary suggestion. 

\section{Discussion}
The data presented above highlight the potential of CGM to deconstruct the metabolic complexity of early dysglycemia. The multiple datapoints generated by CGM during dynamic conditions such as oral glucose loads or meal challenges are amenable to machine learning approaches that generate predictive algorithms capable of identifying distinct metabolic subphenotypes. Phenotypes including insulin resistance, beta-cell dysfunction, and incretin deficiency have implications for clinical risk, precision nutrition, and lifestyle interventions that may decrease risk for progression to diabetes. The ability to identify these metabolic phenotypes at home using CGM has benefits with regard to low patient burden (compared to OGTT in a clinical setting) and high patient appeal.   CGM glucose patterns may also be able to inform precision medical therapy that is targeted towards the predicted physiologic deficit. Future studies should address the stability of glycemic patterns over time and whether the expected shift from an IR pattern to an IS pattern on CGM ensues following interventions that improve insulin sensitivity (e.g., weight loss), so that the CGM could be used longitudinally to evaluate changes in metabolic risk profile in the same individual. In the near future, smartwatch data could be integrated with CGM for metabolic subphenotyping prediction~\citep{Metwally2025, Klonoff2025}, and integrated into conversational artificial intelligence systems to help users interpret results~\citep{Metwally2025, Zhang2025, Mallinar2025}. Thus, with the potential of CGM to improve human health still in its infancy, the future for diabetes and metabolic diseases may be brighter as a result of this relatively new wearable technology.

\section*{ACKNOWLEDGEMENTS:}
The authors thank Lettie McGuire for her tremendous work on the illustrations used in this work.

\section*{DISCLOSURES:}
AAM is an employee of Alphabet and may own stock as part of the standard compensation Package. TM is a member of the scientific advisory board of January AI. MPS a cofounder and scientific advisor of Xthera, Exposomics, Filtricine, Fodsel, iollo, InVu Health, January AI, Marble Therapeutics, Mirvie, Next Thought AI, Orange Street Ventures, Personalis, Protos Biologics, Qbio, RTHM, SensOmics. MPS is a scientific advisor of Abbratech, Applied Cognition, Enovone, Jupiter Therapeutics, M3 Helium, Mitrix, Neuvivo, Onza, Sigil Biosciences, Captify Inc, WndrHLTH, Yuvan Research, Ovul.

% \pagebreak
\bibliography{cgm}

\begin{thebibliography}{71}
\providecommand{\natexlab}[1]{#1}
\providecommand{\url}[1]{\texttt{#1}}
\expandafter\ifx\csname urlstyle\endcsname\relax
  \providecommand{\doi}[1]{doi: #1}\else
  \providecommand{\doi}{doi: \begingroup \urlstyle{rm}\Url}\fi

\bibitem[Abdul-Ghani et~al.(2008)Abdul-Ghani, Abdul-Ghani, Ali, and
  Defronzo]{AbdulGhani2008}
M.~A. Abdul-Ghani, T.~Abdul-Ghani, N.~Ali, and R.~A. Defronzo.
\newblock One-hour plasma glucose concentration and the {Metabolic} {Syndrome}
  identify subjects at high risk for future {Type} 2 {Diabetes}.
\newblock \emph{Diabetes Care}, 31\penalty0 (8):\penalty0 1650--1655, 2008.
\newblock \doi{10.2337/dc08-0225}.
\newblock URL \url{https://doi.org/10.2337/dc08-0225}.

\bibitem[Abdul-Ghani et~al.(2010)Abdul-Ghani, Lyssenko, Tuomi, Defronzo, and
  Groop]{AbdulGhani2010}
M.~A. Abdul-Ghani, V.~Lyssenko, T.~Tuomi, R.~A. Defronzo, and L.~Groop.
\newblock The shape of plasma glucose concentration curve during {OGTT}
  predicts future risk of {Type} 2 {Diabetes}.
\newblock \emph{Diabetes/metabolism Research and Reviews}, 26\penalty0
  (4):\penalty0 280--286, 2010.
\newblock \doi{10.1002/dmrr.1084}.
\newblock URL \url{https://doi.org/10.1002/dmrr.1084}.

\bibitem[{ADA}(2025)]{ADA2025}
{ADA}.
\newblock The american diabetes association releases standards of care in
  diabetes — 2025.
\newblock
  \url{https://diabetes.org/newsroom/press-releases/american-diabetes-association-releases-standards-care-diabetes-2025},
  2025.
\newblock Accessed: October 4, 2025.

\bibitem[Ahlqvist et~al.(2018)Ahlqvist, Storm, Käräjämäki, Martinell,
  Dorkhan, Carlsson, Vikman, and et~al.]{Ahlqvist2018}
E.~Ahlqvist, P.~Storm, A.~Käräjämäki, M.~Martinell, M.~Dorkhan,
  A.~Carlsson, P.~Vikman, and et~al.
\newblock Novel subgroups of adult-onset diabetes and their association with
  outcomes: {A} data-driven cluster analysis of six variables.
\newblock \emph{The Lancet. Diabetes \& Endocrinology}, 6\penalty0
  (5):\penalty0 361--369, 2018.
\newblock \doi{10.1016/S2213-8587(18)30051-2}.
\newblock URL \url{https://doi.org/10.1016/S2213-8587(18)30051-2}.

\bibitem[Alyass et~al.(2015)Alyass, Almgren, Akerlund, Dushoff, Isomaa,
  Nilsson, Tuomi, Lyssenko, Groop, and Meyre]{Alyass2015}
A.~Alyass, P.~Almgren, M.~Akerlund, J.~Dushoff, B.~Isomaa, P.~Nilsson,
  T.~Tuomi, V.~Lyssenko, L.~Groop, and D.~Meyre.
\newblock Modelling of {OGTT} curve identifies 1 {H} plasma glucose level as a
  strong predictor of incident {Type} 2 {Diabetes}: {Results} from two
  prospective cohorts.
\newblock \emph{Diabetologia}, 58\penalty0 (1):\penalty0 80--87, 2015.
\newblock \doi{10.1007/s00125-014-3390-x}.
\newblock URL \url{https://doi.org/10.1007/s00125-014-3390-x}.

\bibitem[Ancoli-Israel et~al.(2003)Ancoli-Israel, Cole, Alessi, Chambers,
  Moorcroft, and Pollak]{AncoliIsrael2003}
S.~Ancoli-Israel, R.~Cole, C.~Alessi, M.~Chambers, W.~Moorcroft, and C.~P.
  Pollak.
\newblock The role of actigraphy in the study of sleep and circadian rhythms.
\newblock \emph{Sleep}, 26\penalty0 (3):\penalty0 342--392, 2003.
\newblock \doi{10.1093/sleep/26.3.342}.
\newblock URL \url{https://doi.org/10.1093/sleep/26.3.342}.

\bibitem[Anjana et~al.(2020)Anjana, Baskar, Nair, Jebarani, Siddiqui, Pradeepa,
  Unnikrishnan, Palmer, Pearson, and Mohan]{Anjana2020}
R.~M. Anjana, V.~Baskar, A.~T.~N. Nair, S.~Jebarani, M.~K. Siddiqui,
  R.~Pradeepa, R.~Unnikrishnan, C.~Palmer, E.~Pearson, and V.~Mohan.
\newblock Novel subgroups of {Type} 2 {Diabetes} and their association with
  microvascular outcomes in an {Asian} {Indian} population: {A} data-driven
  cluster analysis: {The} {INSPIRED} {Study}.
\newblock \emph{BMJ Open Diabetes Research \& Care}, 8\penalty0 (1):\penalty0
  e001506, 2020.
\newblock \doi{10.1136/bmjdrc-2020-001506}.
\newblock URL \url{https://doi.org/10.1136/bmjdrc-2020-001506}.

\bibitem[Asnicar et~al.(2021)Asnicar, Berry, Valdes, Nguyen, Piccinno, Drew,
  Leeming, and et~al.]{Asnicar2021}
F.~Asnicar, S.~E. Berry, A.~M. Valdes, L.~H. Nguyen, G.~Piccinno, D.~A. Drew,
  E.~Leeming, and et~al.
\newblock Microbiome connections with host metabolism and habitual diet from
  1,098 deeply phenotyped individuals.
\newblock \emph{Nature Medicine}, 27\penalty0 (2):\penalty0 321--332, 2021.
\newblock \doi{10.1038/s41591-020-01183-8}.

\bibitem[Bermingham et~al.(2024)Bermingham, Linenberg, Polidori, Asnicar,
  Arrè, Wolf, Badri, and et~al.]{Bermingham2024}
K.~M. Bermingham, I.~Linenberg, L.~Polidori, F.~Asnicar, A.~Arrè, J.~Wolf,
  F.~Badri, and et~al.
\newblock Effects of a personalized nutrition program on cardiometabolic
  health: {A} randomized controlled trial.
\newblock \emph{Nature Medicine}, 30\penalty0 (7):\penalty0 1888--1897, 2024.
\newblock \doi{10.1038/s41591-024-03023-7}.

\bibitem[Berry et~al.(2020)Berry, Valdes, Drew, Asnicar, Mazidi, Wolf,
  Capdevila, and et~al.]{Berry2020}
S.~E. Berry, A.~M. Valdes, D.~A. Drew, F.~Asnicar, M.~Mazidi, J.~Wolf,
  J.~Capdevila, and et~al.
\newblock Human postprandial responses to food and potential for precision
  nutrition.
\newblock \emph{Nature Medicine}, 26\penalty0 (6):\penalty0 964--973, 2020.
\newblock \doi{10.1038/s41591-020-0934-0}.

\bibitem[Bervoets et~al.(2015)Bervoets, Mewis, and Massa]{Bervoets2015}
L.~Bervoets, A.~Mewis, and G.~Massa.
\newblock The shape of the plasma glucose curve during an oral glucose
  tolerance test as an indicator of beta cell function and insulin sensitivity
  in end-pubertal obese girls.
\newblock \emph{Hormone and Metabolic Research}, 47\penalty0 (6):\penalty0
  437--443, 2015.
\newblock \doi{10.1055/s-0034-1395551}.
\newblock URL \url{https://doi.org/10.1055/s-0034-1395551}.

\bibitem[Cavalot et~al.(2011)Cavalot, Pagliarino, Valle, Di~Martino, Bonomo,
  Massucco, Anfossi, and Trovati]{Cavalot2011}
F.~Cavalot, A.~Pagliarino, M.~Valle, L.~Di~Martino, K.~Bonomo, P.~Massucco,
  G.~Anfossi, and M.~Trovati.
\newblock Postprandial blood glucose predicts cardiovascular events and
  all-cause mortality in {Type} 2 {Diabetes} in a 14-year follow-up: {Lessons}
  from the {San} {Luigi} {Gonzaga} {Diabetes} {Study}.
\newblock \emph{Diabetes Care}, 34\penalty0 (10):\penalty0 2237--2243, 2011.
\newblock \doi{10.2337/dc10-2414}.
\newblock URL \url{https://doi.org/10.2337/dc10-2414}.

\bibitem[{CDC}(2024)]{CDC2024}
{CDC}.
\newblock National {Diabetes} {Statistics} {Report}.
\newblock \url{https://www.cdc.gov/diabetes/php/data-research/index.html}, July
  2024.
\newblock Accessed: October 4, 2025.

\bibitem[Chaix et~al.(2014)Chaix, Zarrinpar, Miu, and Panda]{Chaix2014}
A.~Chaix, A.~Zarrinpar, P.~Miu, and S.~Panda.
\newblock Time-restricted feeding is a preventative and therapeutic
  intervention against diverse nutritional challenges.
\newblock \emph{Cell Metabolism}, 20\penalty0 (6):\penalty0 991--1005, 2014.
\newblock \doi{10.1016/j.cmet.2014.11.001}.
\newblock URL \url{https://doi.org/10.1016/j.cmet.2014.11.001}.

\bibitem[Chatterjee et~al.(2018)Chatterjee, Davies, Heller, Speight, Snoek, and
  Khunti]{Chatterjee2018}
S.~Chatterjee, M.~J. Davies, S.~Heller, J.~Speight, F.~J. Snoek, and K.~Khunti.
\newblock Diabetes structured self-management education programmes: {A}
  narrative review and current innovations.
\newblock \emph{The Lancet. Diabetes \& Endocrinology}, 6\penalty0
  (2):\penalty0 130--142, 2018.
\newblock \doi{10.1016/S2213-8587(17)30239-5}.
\newblock URL \url{https://doi.org/10.1016/S2213-8587(17)30239-5}.

\bibitem[Cheng(2015)]{Cheng2015}
C.~K.~L. Cheng.
\newblock The areas under curves ({AUC}) used in diabetes research: {Update}
  view.
\newblock \emph{Int Obesity Diabetes}, 4\penalty0 (1):\penalty0 1--8, 2015.
\newblock URL
  \url{https://www.oatext.com/the-areas-under-curves-auc-used-in-diabetes-research-update-view.php}.

\bibitem[Cheng et~al.(2019)Cheng, Yang, Li, Sun, Qiu, Xu, Ping, Li, and
  Zhang]{Cheng2019}
X.~Cheng, N.~Yang, Y.~Li, Q.~Sun, L.~Qiu, L.~Xu, F.~Ping, W.~Li, and H.~Zhang.
\newblock The shape of the glucose response curve during an oral glucose
  tolerance test heralds beta-cell function in a large {Chinese} population.
\newblock \emph{BMC Endocrine Disorders}, 19\penalty0 (1):\penalty0 114, 2019.
\newblock \doi{10.1186/s12902-019-0446-4}.
\newblock URL \url{https://doi.org/10.1186/s12902-019-0446-4}.

\bibitem[Chung et~al.(2017)Chung, Ha, Onuzuruike, Kasturi, Galvan-De La~Cruz,
  Bingham, Baker, and et~al.]{Chung2017}
S.~T. Chung, J.~Ha, A.~U. Onuzuruike, K.~Kasturi, M.~Galvan-De La~Cruz, B.~A.
  Bingham, R.~L. Baker, and et~al.
\newblock Time to glucose peak during an oral glucose tolerance test identifies
  prediabetes risk.
\newblock \emph{Clinical Endocrinology}, 87\penalty0 (5):\penalty0 484--491,
  2017.
\newblock \doi{10.1111/cen.13416}.
\newblock URL \url{https://doi.org/10.1111/cen.13416}.

\bibitem[Dwibedi et~al.(2022)Dwibedi, Mellergård, Gyllensten, Nilsson,
  Axelsson, Bäckman, Sahlgren, and et~al.]{Dwibedi2022}
C.~Dwibedi, E.~Mellergård, A.~C. Gyllensten, K.~Nilsson, A.~S. Axelsson,
  M.~Bäckman, M.~Sahlgren, and et~al.
\newblock Effect of self-managed lifestyle treatment on glycemic control in
  patients with {Type} 2 {Diabetes}.
\newblock \emph{Npj Digital Medicine}, 5\penalty0 (1):\penalty0 1--13, 2022.
\newblock \doi{10.1038/s41746-022-00627-3}.

\bibitem[Engelgau et~al.(1997)Engelgau, Thompson, Herman, Boyle, Aubert, Kenny,
  Badran, Sous, and Ali]{Engelgau1997}
M.~M. Engelgau, T.~J. Thompson, W.~H. Herman, J.~P. Boyle, R.~E. Aubert, S.~J.
  Kenny, A.~Badran, E.~S. Sous, and M.~A. Ali.
\newblock Comparison of fasting and 2-hour glucose and {HbA1c} levels for
  diagnosing diabetes. {Diagnostic} criteria and performance revisited.
\newblock \emph{Diabetes Care}, 20\penalty0 (5):\penalty0 785--791, 1997.
\newblock \doi{10.2337/diacare.20.5.785}.
\newblock URL \url{https://doi.org/10.2337/diacare.20.5.785}.

\bibitem[Frøslie et~al.(2013)Frøslie, Røislien, Qvigstad, Godang,
  Bollerslev, Voldner, Henriksen, and Veierød]{Froslie2013}
K.~F. Frøslie, J.~Røislien, E.~Qvigstad, K.~Godang, J.~Bollerslev,
  N.~Voldner, T.~Henriksen, and M.~B. Veierød.
\newblock Shape information from glucose curves: {Functional} data analysis
  compared with traditional summary measures.
\newblock \emph{BMC Medical Research Methodology}, 13\penalty0 (1):\penalty0 6,
  2013.
\newblock \doi{10.1186/1471-2288-13-6}.
\newblock URL \url{https://doi.org/10.1186/1471-2288-13-6}.

\bibitem[Gabriel and Zierath(2019)]{Gabriel2019}
B.~M. Gabriel and J.~R. Zierath.
\newblock Circadian rhythms and exercise - re-setting the clock in metabolic
  disease.
\newblock \emph{Nature Reviews. Endocrinology}, 15\penalty0 (4):\penalty0
  197--206, 2019.
\newblock \doi{10.1038/s41574-018-0150-x}.
\newblock URL \url{https://doi.org/10.1038/s41574-018-0150-x}.

\bibitem[Gallwitz(2009)]{Gallwitz2009}
B.~Gallwitz.
\newblock Implications of postprandial glucose and weight control in people
  with {Type} 2 {Diabetes}: {Understanding} and implementing the
  {International} {Diabetes} {Federation} {Guidelines}.
\newblock \emph{Diabetes Care}, 32\penalty0 (Suppl 2):\penalty0 S331--S337,
  2009.
\newblock \doi{10.2337/dc09-S331}.
\newblock URL \url{https://doi.org/10.2337/dc09-S331}.

\bibitem[Hall et~al.(2018)Hall, Perelman, Breschi, Limcaoco, Kellogg,
  McLaughlin, and Snyder]{Hall2018}
H.~Hall, D.~Perelman, A.~Breschi, P.~Limcaoco, R.~Kellogg, T.~McLaughlin, and
  M.~Snyder.
\newblock Glucotypes reveal new patterns of glucose dysregulation.
\newblock \emph{PLoS Biology}, 16\penalty0 (7):\penalty0 e2005143, 2018.
\newblock \doi{10.1371/journal.pbio.2005143}.
\newblock URL \url{https://doi.org/10.1371/journal.pbio.2005143}.

\bibitem[Hanson et~al.(1993)Hanson, Nelson, McCance, Beart, Charles, Pettitt,
  and Knowler]{Hanson1993}
R.~L. Hanson, R.~G. Nelson, D.~R. McCance, J.~A. Beart, M.~A. Charles, D.~J.
  Pettitt, and W.~C. Knowler.
\newblock Comparison of screening tests for non-insulin-dependent diabetes
  mellitus.
\newblock \emph{Archives of Internal Medicine}, 153\penalty0 (18):\penalty0
  2133--2140, 1993.
\newblock URL \url{https://pubmed.ncbi.nlm.nih.gov/8379805/}.

\bibitem[Hulman et~al.(2017)Hulman, Vistisen, Glümer, Bergman, Witte, and
  Færch]{Hulman2017}
A.~Hulman, D.~Vistisen, C.~Glümer, M.~Bergman, D.~R. Witte, and K.~Færch.
\newblock Glucose patterns during an oral glucose tolerance test and
  associations with future diabetes, cardiovascular disease and all-cause
  mortality rate.
\newblock \emph{Diabetologia}, 61\penalty0 (1):\penalty0 101--107, 2017.
\newblock \doi{10.1007/s00125-017-4467-0}.

\bibitem[Hulman et~al.(2018)Hulman, Witte, Vistisen, Balkau, Dekker, Herder,
  Hatunic, Konrad, Færch, and Manco]{Hulman2018}
A.~Hulman, D.~R. Witte, D.~Vistisen, B.~Balkau, J.~M. Dekker, C.~Herder,
  M.~Hatunic, T.~Konrad, K.~Færch, and M.~Manco.
\newblock Pathophysiological characteristics underlying different glucose
  response curves: {A} latent class trajectory analysis from the prospective
  {EGIR-RISC} {Study}.
\newblock \emph{Diabetes Care}, 41\penalty0 (8):\penalty0 1740--1748, 2018.
\newblock \doi{10.2337/dc18-0428}.

\bibitem[Ismail et~al.(2018)Ismail, Xu, Libman, Becker, Marks, Skyler, Palmer,
  and Sosenko]{Ismail2018}
H.~M. Ismail, P.~Xu, I.~M. Libman, D.~J. Becker, J.~B. Marks, J.~S. Skyler,
  J.~P. Palmer, and J.~M. Sosenko.
\newblock The shape of the glucose concentration curve during an oral glucose
  tolerance test predicts risk for {Type} 1 {Diabetes}.
\newblock \emph{Diabetologia}, 61\penalty0 (1):\penalty0 84--92, 2018.
\newblock \doi{10.1007/s00125-017-4453-6}.
\newblock URL \url{https://doi.org/10.1007/s00125-017-4453-6}.

\bibitem[Jagannathan et~al.(2020)Jagannathan, Neves, Dorcely, Chung, Tamura,
  Rhee, and Bergman]{Jagannathan2020}
R.~Jagannathan, J.~S. Neves, B.~Dorcely, S.~T. Chung, K.~Tamura, M.~Rhee, and
  M.~Bergman.
\newblock The oral glucose tolerance test: 100 years later.
\newblock \emph{Diabetes, Metabolic Syndrome and Obesity}, 13:\penalty0
  3787--3805, 2020.
\newblock \doi{10.2147/DMSO.S266062}.

\bibitem[Kaga et~al.(2020)Kaga, Tamura, Takeno, Kakehi, Someya, Funayama,
  Furukawa, and et~al.]{Kaga2020}
H.~Kaga, Y.~Tamura, K.~Takeno, S.~Kakehi, Y.~Someya, T.~Funayama, Y.~Furukawa,
  and et~al.
\newblock Shape of the glucose response curve during an oral glucose tolerance
  test is associated with insulin clearance and muscle insulin sensitivity in
  healthy non-obese men.
\newblock \emph{Journal of Diabetes Investigation}, 11\penalty0 (4):\penalty0
  860--868, 2020.
\newblock \doi{10.1111/jdi.13227}.
\newblock URL \url{https://doi.org/10.1111/jdi.13227}.

\bibitem[Kanauchi et~al.(2005)Kanauchi, Kimura, Kanauchi, and
  Saito]{Kanauchi2005}
M.~Kanauchi, K.~Kimura, K.~Kanauchi, and Y.~Saito.
\newblock Beta-cell function and insulin sensitivity contribute to the shape of
  plasma glucose curve during an oral glucose tolerance test in non-diabetic
  individuals.
\newblock \emph{International Journal of Clinical Practice}, 59\penalty0
  (4):\penalty0 427--431, 2005.
\newblock \doi{10.1111/j.1368-5031.2005.00422.x}.
\newblock URL \url{https://doi.org/10.1111/j.1368-5031.2005.00422.x}.

\bibitem[Karpati et~al.(2018)Karpati, Leventer-Roberts, Feldman, Cohen-Stavi,
  Raz, and Balicer]{Karpati2018}
T.~Karpati, M.~Leventer-Roberts, B.~Feldman, C.~Cohen-Stavi, I.~Raz, and
  R.~Balicer.
\newblock Patient clusters based on {HbA1c} trajectories: {A} step toward
  individualized medicine in {Type} 2 {Diabetes}.
\newblock \emph{PloS One}, 13\penalty0 (11):\penalty0 e0207096, 2018.
\newblock \doi{10.1371/journal.pone.0207096}.
\newblock URL \url{https://doi.org/10.1371/journal.pone.0207096}.

\bibitem[Kasturi et~al.(2019)Kasturi, Onuzuruike, Kunnam, Shomaker, Yanovski,
  and Chung]{Kasturi2019}
K.~Kasturi, A.~U. Onuzuruike, S.~Kunnam, L.~B. Shomaker, J.~A. Yanovski, and
  S.~T. Chung.
\newblock Two- vs one-hour glucose tolerance testing: {Predicting} prediabetes
  in adolescent girls with obesity.
\newblock \emph{Pediatric Diabetes}, 20\penalty0 (2):\penalty0 214--221, 2019.
\newblock \doi{10.1111/pedi.12803}.
\newblock URL \url{https://doi.org/10.1111/pedi.12803}.

\bibitem[Kim et~al.(2012)Kim, Coletta, Mandarino, and Shaibi]{Kim2012}
J.~Y. Kim, D.~K. Coletta, L.~J. Mandarino, and G.~Q. Shaibi.
\newblock Glucose response curve and {Type} 2 {Diabetes} risk in {Latino}
  adolescents.
\newblock \emph{Diabetes Care}, 35\penalty0 (9):\penalty0 1925--1930, 2012.
\newblock \doi{10.2337/dc11-2544}.

\bibitem[Kim et~al.(2016)Kim, Michaliszyn, Nasr, Lee, Tfayli, Hannon, Hughan,
  Bacha, and Arslanian]{Kim2016}
J.~Y. Kim, S.~F. Michaliszyn, A.~Nasr, S.~Lee, H.~Tfayli, T.~Hannon, K.~S.
  Hughan, F.~Bacha, and S.~Arslanian.
\newblock The shape of the glucose response curve during an oral glucose
  tolerance test heralds biomarkers of {Type} 2 {Diabetes} risk in obese youth.
\newblock \emph{Diabetes Care}, 39\penalty0 (8):\penalty0 1431--1439, 2016.
\newblock \doi{10.2337/dc16-0352}.
\newblock URL \url{https://doi.org/10.2337/dc16-0352}.

\bibitem[Klonoff et~al.(2025)Klonoff, Bergenstal, Cengiz, Clements, Espes,
  Espinoza, Kerr, and et~al.]{Klonoff2025}
D.~C. Klonoff, R.~M. Bergenstal, E.~Cengiz, M.~A. Clements, D.~Espes,
  J.~Espinoza, D.~Kerr, and et~al.
\newblock {CGM} data analysis 2.0: {Functional} data pattern recognition and
  artificial intelligence applications.
\newblock \emph{Journal of Diabetes Science and Technology}, page
  19322968251353228, 2025.
\newblock \doi{10.1177/19322968251353228}.
\newblock URL
  \url{https://journals.sagepub.com/doi/abs/10.1177/19322968251353228}.
\newblock Accessed: October 12, 2025.

\bibitem[Knowler et~al.(2002)Knowler, Barrett-Connor, Fowler, Hamman, Lachin,
  Walker, and Nathan]{Knowler2002}
W.~C. Knowler, E.~Barrett-Connor, S.~E. Fowler, R.~F. Hamman, J.~M. Lachin,
  E.~A. Walker, and D.~M. Nathan.
\newblock Reduction in the incidence of {Type} 2 {Diabetes} with lifestyle
  intervention or metformin.
\newblock \emph{The New England Journal of Medicine}, 346\penalty0
  (6):\penalty0 393--403, 2002.
\newblock \doi{10.1056/NEJMoa012512}.
\newblock URL \url{https://doi.org/10.1056/NEJMoa012512}.

\bibitem[Kobayashi et~al.(2025)Kobayashi, Linden-Santangeli, Chan, Toomey,
  Mudaliar, Temprosa, Edelstein, Goyal, Rangamani, and Majithia]{Kobayashi2025}
E.~Kobayashi, N.~J. Linden-Santangeli, N.~Chan, C.~B. Toomey, S.~Mudaliar,
  M.~Temprosa, S.~Edelstein, R.~Goyal, P.~Rangamani, and A.~R. Majithia.
\newblock Longitudinal metabolic trajectories in {Diabetes} {Prevention}
  {Program} participants reveal subgroups with varying micro- and macrovascular
  complication risks.
\newblock \emph{Diabetes Care}, 48\penalty0 (10):\penalty0 1866--1874, 2025.
\newblock \doi{10.2337/dc25-0866}.
\newblock URL \url{https://doi.org/10.2337/dc25-0866}.

\bibitem[Legaard et~al.(2023)Legaard, Lyngbæk, Almdal, Karstoft, Bennetsen,
  Feineis, Nielsen, and et~al.]{Legaard2023}
G.~E. Legaard, M.~P.~P. Lyngbæk, T.~P. Almdal, K.~Karstoft, S.~L. Bennetsen,
  C.~S. Feineis, N.~S. Nielsen, and et~al.
\newblock Effects of different doses of exercise and diet-induced weight loss
  on beta-cell function in {Type} 2 {Diabetes} ({DOSE-EX}): {A} randomized
  clinical trial.
\newblock \emph{Nature Metabolism}, 5\penalty0 (5):\penalty0 880--895, 2023.
\newblock \doi{10.1038/s42255-023-00796-y}.

\bibitem[Mahajan et~al.(2022)Mahajan, Spracklen, Zhang, Ng, Petty, Kitajima,
  Yu, and et~al.]{Mahajan2022}
A.~Mahajan, C.~N. Spracklen, W.~Zhang, M.~C.~Y. Ng, L.~E. Petty, H.~Kitajima,
  G.~Z. Yu, and et~al.
\newblock Multi-ancestry genetic study of {Type} 2 {Diabetes} highlights the
  power of diverse populations for discovery and translation.
\newblock \emph{Nature Genetics}, 54\penalty0 (5):\penalty0 560--572, 2022.
\newblock \doi{10.1038/s41588-022-01058-3}.

\bibitem[Mallinar et~al.(2025)Mallinar, Heydari, Liu, Faranesh, Winslow,
  Hammerquist, Graef, and et~al.]{Mallinar2025}
N.~Mallinar, A.~A. Heydari, X.~Liu, A.~Z. Faranesh, B.~Winslow, N.~Hammerquist,
  B.~Graef, and et~al.
\newblock A scalable framework for evaluating health language models.
\newblock \emph{arXiv}, 2025.
\newblock URL \url{http://arxiv.org/abs/2503.23339}.

\bibitem[Manco et~al.(2017)Manco, Nolfe, Pataky, Monti, Porcellati, Gabriel,
  Mitrakou, and Mingrone]{Manco2017}
M.~Manco, G.~Nolfe, Z.~Pataky, L.~Monti, F.~Porcellati, R.~Gabriel,
  A.~Mitrakou, and G.~Mingrone.
\newblock Shape of the {OGTT} glucose curve and risk of impaired glucose
  metabolism in the {EGIR-RISC} cohort.
\newblock \emph{Metabolism: Clinical and Experimental}, 70:\penalty0 30--40,
  2017.
\newblock \doi{10.1016/j.metabol.2017.02.007}.
\newblock URL \url{https://doi.org/10.1016/j.metabol.2017.02.007}.

\bibitem[Matsuda and DeFronzo(1999)]{Matsuda1999}
M.~Matsuda and R.~A. DeFronzo.
\newblock Insulin sensitivity indices obtained from oral glucose tolerance
  testing: {Comparison} with the euglycemic insulin clamp.
\newblock \emph{Diabetes Care}, 22\penalty0 (9):\penalty0 1462--1470, 1999.
\newblock \doi{10.2337/diacare.22.9.1462}.

\bibitem[Matthews et~al.(1985)Matthews, Hosker, Rudenski, Naylor, Treacher, and
  Turner]{Matthews1985}
D.~R. Matthews, J.~P. Hosker, A.~S. Rudenski, B.~A. Naylor, D.~F. Treacher, and
  R.~C. Turner.
\newblock Homeostasis model assessment: {Insulin} resistance and beta-cell
  function from fasting plasma glucose and insulin concentrations in man.
\newblock \emph{Diabetologia}, 28\penalty0 (7):\penalty0 412--419, 1985.
\newblock \doi{10.1007/BF00280883}.

\bibitem[McHill et~al.(2017)McHill, Phillips, Czeisler, Keating, Yee, Barger,
  Garaulet, Scheer, and Klerman]{McHill2017}
A.~W. McHill, A.~J. Phillips, C.~A. Czeisler, L.~Keating, K.~Yee, L.~K. Barger,
  M.~Garaulet, F.~A. Scheer, and E.~B. Klerman.
\newblock Later circadian timing of food intake is associated with increased
  body fat.
\newblock \emph{The American Journal of Clinical Nutrition}, 106\penalty0
  (5):\penalty0 1213--1219, 2017.
\newblock \doi{10.3945/ajcn.117.161588}.
\newblock URL \url{https://doi.org/10.3945/ajcn.117.161588}.

\bibitem[Metwally et~al.(2024)Metwally, Perelman, Park, Wu, Jha, Sharp, Celli,
  and et~al.]{Metwally2024}
A.~A. Metwally, D.~Perelman, H.~Park, Y.~Wu, A.~Jha, S.~Sharp, A.~Celli, and
  et~al.
\newblock Prediction of metabolic subphenotypes of {Type} 2 {Diabetes} via
  continuous glucose monitoring and machine learning.
\newblock \emph{Nature Biomedical Engineering}, 9\penalty0 (8):\penalty0
  1222--1239, 2024.
\newblock \doi{10.1038/s41551-024-01217-z}.

\bibitem[Metwally et~al.(2025)Metwally, Heydari, McDuff, Solot, Esmaeilpour,
  Faranesh, Zhou, and et~al.]{Metwally2025}
A.~A. Metwally, A.~A. Heydari, D.~McDuff, A.~Solot, Z.~Esmaeilpour, A.~Z.
  Faranesh, M.~Zhou, and et~al.
\newblock Insulin resistance prediction from wearables and routine blood
  biomarkers.
\newblock \emph{arXiv}, 2025.
\newblock URL \url{http://arxiv.org/abs/2505.03784}.

\bibitem[Nair et~al.(2022)Nair, Wesolowska-Andersen, Brorsson, Rajendrakumar,
  Gan, Dawed, and et~al.]{Nair2022}
A.~T.~N. Nair, A.~Wesolowska-Andersen, C.~Brorsson, A.~L. Rajendrakumar,
  S.~Gan, A.~Y. Dawed, and et~al.
\newblock Heterogeneity in phenotype, disease progression and drug response in
  {Type} 2 {Diabetes}.
\newblock \emph{Nature Medicine}, 28\penalty0 (5):\penalty0 982--988, 2022.
\newblock \doi{10.1038/s41591-022-01799-5}.

\bibitem[Nakamura et~al.(2021)Nakamura, Tajiri, Hatamoto, Ando, Shimoda, and
  Yoshimura]{Nakamura2021}
K.~Nakamura, E.~Tajiri, Y.~Hatamoto, T.~Ando, S.~Shimoda, and E.~Yoshimura.
\newblock Eating dinner early improves 24-h blood glucose levels and boosts
  lipid metabolism after breakfast the next day: {A} randomized cross-over
  trial.
\newblock \emph{Nutrients}, 13\penalty0 (7):\penalty0 2424, 2021.
\newblock \doi{10.3390/nu13072424}.
\newblock URL \url{https://doi.org/10.3390/nu13072424}.

\bibitem[Nauck et~al.(1986)Nauck, Stöckmann, Ebert, and
  Creutzfeldt]{Nauck1986}
M.~Nauck, F.~Stöckmann, R.~Ebert, and W.~Creutzfeldt.
\newblock Reduced incretin effect in {Type} 2 (non-insulin-dependent)
  {Diabetes}.
\newblock \emph{Diabetologia}, 29\penalty0 (1):\penalty0 46--52, 1986.
\newblock \doi{10.1007/BF00400186}.

\bibitem[Nôga et~al.(2024)Nôga, Meth, Pacheco, Tan, Cedernaes, van Egmond,
  Xue, and Benedict]{Noga2024}
D.~A. Nôga, E.~d. M. e.~S. Meth, A.~P. Pacheco, X.~Tan, J.~Cedernaes, L.~T.
  van Egmond, P.~Xue, and C.~Benedict.
\newblock Habitual short sleep duration, diet, and development of {Type} 2
  {Diabetes} in adults.
\newblock \emph{JAMA Network Open}, 7\penalty0 (3):\penalty0 e241147, 2024.
\newblock \doi{10.1001/jamanetworkopen.2024.1147}.

\bibitem[Ohkawara et~al.(2011)Ohkawara, Oshima, Hikihara, Ishikawa-Takata,
  Tabata, and Tanaka]{Ohkawara2011}
K.~Ohkawara, Y.~Oshima, Y.~Hikihara, K.~Ishikawa-Takata, I.~Tabata, and
  S.~Tanaka.
\newblock Real-time estimation of daily physical activity intensity by a
  triaxial accelerometer and a gravity-removal classification algorithm.
\newblock \emph{The British Journal of Nutrition}, 105\penalty0 (11):\penalty0
  1681--1691, 2011.
\newblock \doi{10.1017/S0007114510005441}.
\newblock URL \url{https://doi.org/10.1017/S0007114510005441}.

\bibitem[Pan et~al.(2011)Pan, Schernhammer, Sun, and Hu]{Pan2011}
A.~Pan, E.~S. Schernhammer, Q.~Sun, and F.~B. Hu.
\newblock Rotating night shift work and risk of {Type} 2 {Diabetes}: {Two}
  prospective cohort studies in women.
\newblock \emph{PLoS Medicine}, 8\penalty0 (12):\penalty0 e1001141, 2011.
\newblock \doi{10.1371/journal.pmed.1001141}.
\newblock URL \url{https://doi.org/10.1371/journal.pmed.1001141}.

\bibitem[Panda(2016)]{Panda2016}
S.~Panda.
\newblock Circadian physiology of metabolism.
\newblock \emph{Science (New York, N.Y.)}, 354\penalty0 (6315):\penalty0
  1008--1015, 2016.
\newblock \doi{10.1126/science.aah4967}.
\newblock URL \url{https://doi.org/10.1126/science.aah4967}.

\bibitem[Park et~al.(2025)Park, Metwally, Delfarah, Wu, Perelman, Mayer,
  McGinity, and et~al.]{Park2025}
H.~Park, A.~A. Metwally, A.~Delfarah, Y.~Wu, D.~Perelman, C.~Mayer,
  C.~McGinity, and et~al.
\newblock High-resolution lifestyle profiling and metabolic subphenotypes of
  {Type} 2 {Diabetes}.
\newblock \emph{Npj Digital Medicine}, 8\penalty0 (1):\penalty0 1--14, 2025.
\newblock \doi{10.1038/s41746-024-01188-5}.

\bibitem[Pavlou et~al.(2023)Pavlou, Cienfuegos, Lin, Ezpeleta, Ready, Corapi,
  Wu, and et~al.]{Pavlou2023}
V.~Pavlou, S.~Cienfuegos, S.~Lin, M.~Ezpeleta, K.~Ready, S.~Corapi, J.~Wu, and
  et~al.
\newblock Effect of time-restricted eating on weight loss in adults with {Type}
  2 {Diabetes}: {A} randomized clinical trial.
\newblock \emph{JAMA Network Open}, 6\penalty0 (10):\penalty0 e2339337, 2023.
\newblock \doi{10.1001/jamanetworkopen.2023.39337}.
\newblock URL \url{https://doi.org/10.1001/jamanetworkopen.2023.39337}.

\bibitem[Qu et~al.(2022)Qu, Chen, Chen, and Zhang]{Qu2022}
X.~Qu, K.~Chen, J.~Chen, and J.~Zhang.
\newblock Trends in adherence to recommended physical activity and its effects
  on cardiometabolic markers in {US} adults with pre-diabetes.
\newblock \emph{BMJ Open Diabetes Research \& Care}, 10\penalty0 (5):\penalty0
  e002981, 2022.
\newblock \doi{10.1136/bmjdrc-2022-002981}.
\newblock URL \url{https://doi.org/10.1136/bmjdrc-2022-002981}.

\bibitem[{Report of the Expert Committee on the Diagnosis and Classification of
  Diabetes Mellitus}(1997)]{ExpertCommittee1997}
{Report of the Expert Committee on the Diagnosis and Classification of Diabetes
  Mellitus}.
\newblock Report of the expert committee on the diagnosis and classification of
  diabetes mellitus.
\newblock \emph{Diabetes Care}, 20\penalty0 (7):\penalty0 1183--1197, 1997.
\newblock \doi{10.2337/diacare.20.7.1183}.
\newblock URL \url{https://doi.org/10.2337/diacare.20.7.1183}.

\bibitem[Scheer et~al.(2009)Scheer, Hilton, Mantzoros, and Shea]{Scheer2009}
F.~A. Scheer, M.~F. Hilton, C.~S. Mantzoros, and S.~A. Shea.
\newblock Adverse metabolic and cardiovascular consequences of circadian
  misalignment.
\newblock \emph{Proceedings of the National Academy of Sciences of the United
  States of America}, 106\penalty0 (11):\penalty0 4453--4458, 2009.
\newblock \doi{10.1073/pnas.0808180106}.
\newblock URL \url{https://doi.org/10.1073/pnas.0808180106}.

\bibitem[Shajari et~al.(2023)Shajari, Kuruvinashetti, Komeili, and
  Sundararaj]{Shajari2023}
S.~Shajari, K.~Kuruvinashetti, A.~Komeili, and U.~Sundararaj.
\newblock The emergence of {AI-Based} wearable sensors for digital health
  technology: {A} review.
\newblock \emph{Sensors (Basel, Switzerland)}, 23\penalty0 (23):\penalty0 9498,
  2023.
\newblock \doi{10.3390/s23239498}.
\newblock URL \url{https://doi.org/10.3390/s23239498}.

\bibitem[Shen et~al.(2023)Shen, Ma, Wu, Liu, Chen, and Yang]{Shen2023}
B.~Shen, C.~Ma, G.~Wu, H.~Liu, L.~Chen, and G.~Yang.
\newblock Effects of exercise on circadian rhythms in humans.
\newblock \emph{Frontiers in Pharmacology}, 14:\penalty0 1282357, October 2023.
\newblock \doi{10.3389/fphar.2023.1282357}.
\newblock URL \url{https://doi.org/10.3389/fphar.2023.1282357}.

\bibitem[Shen et~al.(2025)Shen, Li, Gou, Liang, Zhong, Xiao, Shi, and
  et~al.]{Shen2025}
L.~Shen, B.-Y. Li, W.~Gou, X.~Liang, H.~Zhong, C.~Xiao, R.~Shi, and et~al.
\newblock Trajectories of sleep duration, sleep onset timing, and continuous
  glucose monitoring in adults.
\newblock \emph{JAMA Network Open}, 8\penalty0 (3):\penalty0 e250114, 2025.
\newblock \doi{10.1001/jamanetworkopen.2025.0114}.

\bibitem[Tahara and Shibata(2013)]{Tahara2013}
Y.~Tahara and S.~Shibata.
\newblock Chronobiology and nutrition.
\newblock \emph{Neuroscience}, 253:\penalty0 78--88, December 2013.
\newblock \doi{10.1016/j.neuroscience.2013.08.049}.
\newblock URL \url{https://doi.org/10.1016/j.neuroscience.2013.08.049}.

\bibitem[Teymoori et~al.(2023)Teymoori, Jahromi, Ahmadirad, Daftari, Mokhtari,
  Farhadnejad, Mirmiran, and Azizi]{Teymoori2023}
F.~Teymoori, M.~K. Jahromi, H.~Ahmadirad, G.~Daftari, E.~Mokhtari,
  H.~Farhadnejad, P.~Mirmiran, and F.~Azizi.
\newblock The association of dietary and lifestyle indices for insulin
  resistance with the risk of cardiometabolic diseases among {Iranian} adults.
\newblock \emph{Scientific Reports}, 13\penalty0 (1):\penalty0 1--10, 2023.
\newblock \doi{10.1038/s41598-023-45542-8}.

\bibitem[Tänczer et~al.(2020)Tänczer, Svébis, Domján, Horváth, and
  Tabák]{Tanczer2020}
T.~Tänczer, M.~M. Svébis, B.~Domján, V.~J. Horváth, and A.~G. Tabák.
\newblock The effect of prior gestational diabetes on the shape of the glucose
  response curve during an oral glucose tolerance test 3 years after delivery.
\newblock \emph{Journal of Diabetes Research}, 2020:\penalty0 4315806, March
  2020.
\newblock \doi{10.1155/2020/4315806}.
\newblock URL \url{https://doi.org/10.1155/2020/4315806}.

\bibitem[Udler et~al.(2019)Udler, McCarthy, Florez, and Mahajan]{Udler2019}
M.~S. Udler, M.~I. McCarthy, J.~C. Florez, and A.~Mahajan.
\newblock Genetic risk scores for diabetes diagnosis and precision medicine.
\newblock \emph{Endocrine Reviews}, 40\penalty0 (6):\penalty0 1500--1520, 2019.
\newblock \doi{10.1210/er.2019-00088}.
\newblock URL \url{https://doi.org/10.1210/er.2019-00088}.

\bibitem[Wehrens et~al.(2017)Wehrens, Christou, Isherwood, Middleton, Gibbs,
  Archer, Skene, and Johnston]{Wehrens2017}
S.~M.~T. Wehrens, S.~Christou, C.~Isherwood, B.~Middleton, M.~A. Gibbs, S.~N.
  Archer, D.~J. Skene, and J.~D. Johnston.
\newblock Meal timing regulates the human circadian system.
\newblock \emph{Current Biology : CB}, 27\penalty0 (12):\penalty0
  1768--1775.e3, 2017.
\newblock \doi{10.1016/j.cub.2017.04.059}.
\newblock URL \url{https://doi.org/10.1016/j.cub.2017.04.059}.

\bibitem[Wu et~al.(2025)Wu, Ehlert, Metwally, Perelman, Park, Brooks, Abbasi,
  and et~al.]{Wu2025}
Y.~Wu, B.~Ehlert, A.~A. Metwally, D.~Perelman, H.~Park, A.~W. Brooks,
  F.~Abbasi, and et~al.
\newblock Individual variations in glycemic responses to carbohydrates and
  underlying metabolic physiology.
\newblock \emph{Nature Medicine}, 31\penalty0 (7):\penalty0 2232--2243, 2025.
\newblock \doi{10.1038/s41591-025-03612-y}.

\bibitem[Zambotti et~al.(2019)Zambotti, Cellini, Goldstone, Colrain, and
  Baker]{Zambotti2019}
M.~d. Zambotti, N.~Cellini, A.~Goldstone, I.~M. Colrain, and F.~C. Baker.
\newblock Wearable sleep technology in clinical and research settings.
\newblock \emph{Medicine and Science in Sports and Exercise}, 51\penalty0
  (7):\penalty0 1538--1554, 2019.
\newblock \doi{10.1249/MSS.0000000000001947}.
\newblock URL \url{https://doi.org/10.1249/MSS.0000000000001947}.

\bibitem[Zeevi et~al.(2015)Zeevi, Korem, Zmora, Israeli, Rothschild,
  Weinberger, Ben-Yacov, and et~al.]{Zeevi2015}
D.~Zeevi, T.~Korem, N.~Zmora, D.~Israeli, D.~Rothschild, A.~Weinberger,
  O.~Ben-Yacov, and et~al.
\newblock Personalized nutrition by prediction of glycemic responses.
\newblock \emph{Cell}, 163\penalty0 (5):\penalty0 1079--1094, 2015.
\newblock \doi{10.1016/j.cell.2015.11.001}.
\newblock URL \url{https://doi.org/10.1016/j.cell.2015.11.001}.

\bibitem[Zhang et~al.(2025)Zhang, Ayush, Qiao, Heydari, Narayanswamy, Xu,
  Metwally, and et~al.]{Zhang2025}
Y.~Zhang, K.~Ayush, S.~Qiao, A.~A. Heydari, G.~Narayanswamy, M.~A. Xu, A.~A.
  Metwally, and et~al.
\newblock {SensorLM}: Learning the language of wearable sensors.
\newblock \emph{arXiv}, 2025.
\newblock URL \url{http://arxiv.org/abs/2506.09108}.

\end{thebibliography}

\renewcommand{\thesubsection}{Supplementary Table S\arabic{subsection}}
\setcounter{subsection}{0} % Reset counter

% \section*{Supplementary Tables}
% \input{supplementary/s1}
% \input{supplementary/s2}
% \input{supplementary/s3}
% \input{supplementary/s4}
% \input{supplementary/s5}
% \input{supplementary/s6}
% \input{supplementary/s7}
% \input{supplementary/s8}
% \input{supplementary/s9}
% \input{supplementary/s10}
% \input{supplementary/s11}

\end{document}